# The Development of a Comprehensive Spanish Dictionary for Phonetic and Lexical Tagging in Socio-phonetic Research (ESPADA)

**Simon Gonzalez**
The Australian National University
Canberra, Australian Capital Territory, Australia
u1037706@anu.edu.au

**Abstract**
Pronunciation dictionaries are an important component in the process of speech forced alignment. The accuracy of these dictionaries has a strong effect on the aligned speech data since they help the mapping between orthographic transcriptions and acoustic signals. In this paper, I present the creation of a comprehensive pronunciation dictionary in Spanish (ESPADA) that can be used in most of the dialect variants of Spanish data. Current dictionaries focus on specific regional variants, but with the flexible nature of our tool, it can be readily applied to capture the most common phonetic differences across major dialectal variants. We propose improvements to current pronunciation dictionaries as well as mapping other relevant annotations such as morphological and lexical information. In terms of size, it is currently the most complete dictionary with more than 628,000 entries, representing words from 16 countries. All entries come with their corresponding pronunciations, morphological and lexical tagging, and other relevant information for phonetic analysis: stress patterns, phonotactics, IPA transcriptions, and more. This aims to equip socio-phonetic researchers with a complete open-source tool that enhances dialectal research within socio-phonetic frameworks in the Spanish language.

## 1. Introduction

Within current research frameworks in socio-phonetics, workflows are becoming more and more complex due to the amount of data to be processed and the specialisation that the field is experiencing. This has been strongly influenced by the rapid advances in speech processing technologies readily available to be used. One area that has experienced a great level of specialisation is the forced alignment of natural speech data (Bailey 2016; Fruehwald, 2014), which is now widely used in socio-phonetic research (cf. DiCanio et al, 2012; Gonzalez et al, 2018; Strunk et al, 2014). The goal of the forced alignment process is to create time-aligned segmentations at the phonemic level, following acoustic parameter transitions between units, and derived from orthographic transcriptions at the utterance level (Fromont & Hay, 2012; Gonzalez et al., 2018; Kisler et al, 2017; McAuliffe et al., 2017; Reddy & Stanford, 2015). In this sense, the accuracy of the alignment strongly depends on the level of accuracy of the transcription, which works best when it is transcribed annotating the closest to the spoken speech.

An important component of the forced alignment process is the pronunciation dictionary. This stores all the phonemic representation of the segments that will be aligned in the speech data. In the case of English, there are publicly available dictionaries that represent English phonemes in different ways. One well-known available tool in English is the CMU dictionary (CMU, 2016), which uses an encoding from the ARPABET system, representing vowels with two uppercase symbols (e.g. IPA /e/=ARPABET 'EH'[1]), and consonants with one or two uppercase symbols (e.g. /d/='D'; /h/='HH'). Another one is the Disc option within the CELEX dictionary (Baayen el al., 1995), which uses one letter or number to represent a phoneme (e.g. /eɪ/ = '1'; /tʃ/ = 'J'). After establishing the segments, the next step in the forced alignment process is to annotate all available words in the dictionary, which is also referred to as g2p (grapheme-to-phoneme conversion), and it is the process of converting orthographic text into its corresponding phonological transcription.

### 1.1 Challenges in Pronunciation Dictionaries

Here we present four of the relevant challenges when deciding how to annotate words. The first challenge is the type of phonemic representation, whether following more orthographic-oriented conventions, as the CMU, or more IPA-oriented conventions. For example, the word "cat" has three phonological segments: starting with a consonant, followed by a vowel, then ending in another consonant. This can be represented using the phonemic representation /kat/, but we can also use 'KAT'. This last use has been preferred because of computational practicalities, due to having simpler encodings than IPA symbols.

The second challenge is the question on what are the phonological elements that we represent in the annotation. One of them is stress marking. In languages where the same vowel can be in stressed and unstressed syllables, as in Russian and Spanish, for example, one question is whether we represent stress in the annotation or not. In a study on Russian (Gnevsheva et al., 2020), it was found that when the annotation reflected the stress differences in words, the accuracy of the forced alignment improved. Results like this are in line with the selection by some forced aligners to use dictionaries that represent stressed and unstressed syllables for their words.

When it comes to annotating socio-phonetic variation, the challenges become more pronounced. This is especially crucial when deciding the type of annotation since inaccuracies, or errors, in the forced alignment output must be traced back to the real cause: either inaccuracies in the forced alignment algorithm, or mismatches between the speech signal and the transcription. For example, if we want to annotate the word "car", we can either annotate it to reflect rhotic variants, such as American English, where the annotation would be 'kAr' [kaɹ], or it can be annotated to represent non-rhotic variants, such as Australian English 'kA' [kɐː]. This has been approached in three different ways. In one approach, researchers used a rhotic model on a non-rhotic accent, and then did the corresponding corrections post hoc after the alignment (MacKenzie & Turton, 2020). In other approaches, the acoustic model of a non-rhotics variant was used for another non-rhotic variant, as in Fromont & Watson (2016), where the model from British English was used to forced align New Zealand English. A final approach is when the acoustic model is trained on the non-rhotic variant itself, as in Gonzalez et al. (2020), where they trained the acoustic model on the same data, then used this model to force align the data in Australian English.

---

[1] Across this paper, IPA transcriptions will follow the conventional // for phonemes and [] for allophonic transcription. Orthographic spellings will be represented with double quotation marks " ", and pronunciation dictionary entries with single quotation marks ' '.

A final challenge is related to the stage following the forced alignment process. The forced-aligned data is a preparation process to then analyse it in the light of socio-phonetic questions, which can be on specific segments (vowels and consonants), but also on higher levels, such as morphology, syntax or suprasegmentals (e.g., prosody). This means that the level of annotation does not stop at the pronunciation dictionary, but also has an impact at the word and syntactical levels. It is common practice to do this work after the alignment by doing a new wrangling of the data, using other resources available that map aligned words with their corresponding lexical information, such as word type and part of speech classification. However, when trying to map outcome data to new sources, and using new tools, results are prompt to have more errors in the final product of the wrangling process.

### 1.2 Spanish Pronunciation Dictionaries

All these challenges are latent for any language that can be forced aligned, and in this paper, we focus on Spanish and its variants. Here, we assess these factors in relation to Spanish research, and we also look at the current advancements on pronunciation dictionaries. We discuss the areas of improvement and present the results of the development of a resource that aims to contribute to the work on forced alignment (and further) on Spanish socio-phonetic research. The aim is to develop a tool that can be used for two stages: the first one is for the preparation of data for forced alignment, and the second one is for lexical and grammatical tagging (POS) of the forced aligned output. We expect that such a tool can facilitate the transition stage from forced alignment to further analysis without loss of any data and the avoidance of mapping inconsistencies.

## 2. Related Work

In this socio-phonetic revolution, many technologies and resources have been developed across different languages to achieve great accuracy from annotated dictionaries. A seminal work on this type of annotations has been done for CELEX (Baayen el al., 1995). It contains information on four levels:

- Orthography: shows spelling variations for all corresponding words.
- Phonology: shows corresponding phonological transcriptions, and corresponding variations in pronunciation. It also includes syllable information such as syllable structure and stress types, e.g. primary stress.
- Morphology: shows the derivational and compositional structures for all words, showing their inflectional paradigms.
- Syntax: shows the word class, and all word class-specific subcategorizations, with their corresponding argument structures.

This is an extensive work on linguistic annotation available for English, Dutch, and German. To our knowledge, there is not a version for Spanish available in CELEX.

In terms of pronunciation dictionaries used in forced alignment, there have been great advances in both English and Spanish. For English, state of the art available aligners includes MFA (McAuliffe et al., 2017), MAUS (Kisler et al, 2017), FAVE (Rosenfelder et al., 2014), LaBB-CAT (Fromont & Hay, 2012), and DARLA (Reddy & Stanford, 2015). For Spanish, MFA and faseAlign (Wilbanks, 2021) are two widely used aligners[2]. In this paper, we assess only the phonetic dictionaries available for each of these. We do not evaluate their accuracy[3].

### 2.1 TalnUPF

MFA has two available dictionaries. The first one is the *TalnUPF Spanish IPA* dictionary, and its annotations use IPA symbols, showing stress information. A sample[4] of the dictionary is shown below.

| Word | Annotated pronunciation |
|---|---|
| *matamoros* | m a t a m ˈo ɾ o s |
| *torreón* | t o r e ˈo n |
| *campeche* | k a m p ˈe tʃ e |
| *zúñiga* | θ ˈu ɲ i ɣ a |
| *guasave* | g w a s ˈa β e |
| *allende* | a j ˈe n d e |

Table 1: Sample pronunciation dictionary from *TalnUPF IPA*

The second dictionary available is the *TalnUPF Spanish gpA* dictionary. Contrary to the previous one, this uses alphabetic letters (similar practice as in the CMU dictionary) and not IPA symbols. An important observation is that there is no stress specification in this dictionary, contrary to the IPA version. In addition, like the IPA version, it makes a distinction between monophthongs and semivowels, as seen in the sample below, where the letter "u" in the word "guasave" is represented with the semivowel 'w'. This is compared to the vowel /u/, that is represented as 'u' in the word "zúñiga".

---

[2] Other aligners used for Spanish data include EasyAlign (Goldman, 2011), LaBB-CAT (Fromont & Hay, 2012), and PraatAlign (Lubbers & Torreira, 2016).

[3] Gonzalez, S., Grama, J. and Travis, C.E. (2020). Comparing the performance of forced aligners used in sociophonetic research. *Linguistics Vanguard*, vol. 6, no. 1

[4] These words in the dictionary samples were chosen to be compared to the ones given by Wilbanks (2021) for faseAlign.

| Word | Annotated pronunciation |
|---|---|
| *matamoros* | m a t a m o r f o s |
| *torreón* | t o r e o n |
| *campeche* | k a m p e t S e |
| *zúñiga* | T u n~ i G a |
| *guasave* | g w a s a V e |
| *allende* | a L e n d e |

Table 2: Sample pronunciation dictionary from *TalnUPF gpA*

In terms of individual segments, the letter "ñ" uses two separate symbols: 'n~', which is the palatal nasal /ɲ/. The dictionary also makes the correct distinction between single and complex trills in Spanish: simple (/ɾ/='rf') and complex (/r/='r'). Another distinction is the alveolar /l/ using the lowercase 'l', and the upper case 'L' to represent the palatal lateral /ʎ/ or the palatal fricative /ʝ/, depending on the variant. The third characteristic is that the palato-alveolar affricate /tʃ/ is represented with the two letters 'tS'. Since Spanish voiced stops /b, d, g/ are lenited in intervocalic positions ([β, ð, ɣ], respectively), the dictionary accurately makes the distinction between intervocalic contexts and non-intervocalic, for example, "vaca"='baka', "abeja"='aVexa'. This is the same for the other voiced plosives: 'b' and 'V', 'd' and 'D', and 'g' and 'G'. This distinction is also implemented in the IPA version of the same dictionary. In terms of Spanish variants, since it is based on Castilian Spanish, it represents the labiodental voiceless /θ/='T' and the velar voiceless fricative /x/='x', as in the word "jarrazo"= 'x a r a T o'.

## 2.2 faseAlign

The dictionary used in faseAlign follows a similar structure to the *TalnUPF Spanish gpA* dictionary. The similarity is that there is no stress specification in the annotation, but it differs from the TalnUPF in that it does not make a distinction between monophthongs and semivowels, as seen in the words "guasave" and "zúñiga", both of which use the same phonemic representation 'u' for "u".

| Word | Annotated pronunciation |
|---|---|
| *matamoros* | m a t a m o r o s |
| *torreón* | t o R e o n |
| *campeche* | k a m p e CH e |
| *zúñiga* | s u NY i g a |
| *guasave* | g u a s a b e |
| *allende* | a y e n d e |

Table 3: Sample pronunciation dictionary from *faseAlign*

Another major difference is that it does not make the intervocalic distinction for voiced stops, this is, it gives the same representation regardless of their phonological context. In this sense, faseAlign gives a more phonemic representation whereas TalnUPD focuses on a more phonetic one. In terms of the other phonemes, it represents complex trills with an upper case 'R' and the single one with the lower case 'r'. The palato-alveolar affricate is 'CH', the palatal nasal 'NY', and palatal lateral 'y'. Finally, in terms of Spanish variants, faseAlign targets Latin-American dialects where there is no /θ/ and /x/. In the case of /θ/, most of the variants realise it the same as /s/ (called *seseo*, where /θ/ and /s/ merge to /s/), and /x/ is realised as /h/.

## 3. Assessment on Current Tools

When considering that the stage of forced alignment is a transitional stage to prepare the data for phonetic analysis, we assess the available tools in relation to an overall view. It is very important to note that these resources are not necessary lacking for the tasks they carry out, but rather we propose here to improve phonetic annotations that can be easily adapted to most Spanish variants and combine these with other linguistic annotations.

In terms of the lexical and grammatical tagging, CELEX offers great functionality. However, as observed before, it is not available for Spanish. We therefore propose to create an annotated dataset of Spanish words that have the following six features available:

- Orthography: showing spelling variations for all corresponding words based on dialectal differences.
- Phonology: showing corresponding phonological transcriptions, and corresponding variations in pronunciation. The variants can be modified as needed by the users. This flexibility allows us to adapt the dictionary to a virtually infinite number of dialectal phenomena.
- Syllable divisions: showing syllabic information such as syllable structure and stress types.
- Stress: showing the stress pattern of words.
- Morphology: showing the derivational and compositional structures for all words.
- Syntax: showing the word function in the sentence.

The aim of the dictionary is to be both comprehensive and flexible to be used on any documented Spanish variant. In the table below, we show the functions available in the three dictionaries discussed, and the areas where we aim to contribute with this tool.

| Tool | Different Variants | Alphabetic annotations | Semi-vowels distinction | Stress annotations | Voiced Stops distinction |
|---|---|---|---|---|---|
| faseAlign | Yes | Yes | No | No | No |
| Taln gpA | No | Yes | Yes | No | Yes |
| Taln IPA | No | No | Yes | Yes | Yes |

Table 4: Feature comparison for the three pronunciation dictionaries for Spanish data.

## 4. Aims of the Paper

By implementing these improvements in the annotations, our final product includes both phonetic/phonological layers, as well as grammatical/lexical layers, compared to CELEX. We have called it *ESPADA ([E]Spanish*

*[A]nnotation [Da]ta)*. The first advantage is that all these will be available in one resource, which minimises inaccuracies and inconsistencies in data mapping, and offers users the opportunity to change between all available dialectal options without recreating new pronunciation dictionaries.

Based on a preliminary literature review on major dialectal variants, we identified major phonological contexts which capture regional variants, and they are described as follows[5]:

- Stress: choose annotations with or without stress information.
- Voiced Stops Lenition: choose whether Voiced Stops are only represented in their phonological form /b, d, g/ or with their lenited intervocalic counterparts [β, ð, ɣ].
- Semivowels: specify whether semivowels are represented with only their phonological forms /i, u/, or with their phonetic representations [j, w].
- /ʝ/: choose variants where /ʝ/ is realised as /ʎ/.
- /θ/: choose variants where /θ/ and /s/ are distinctive, or where they merge into /s/.
- /x/: choose variants where /x/ is realised as /h/.
- /s/: choose /h/ in contexts where the /s/ is debuccalized (Morris, 2000). In this case, we have chosen the phonological context where this has been most commonly observed: in post-nuclear position, such as "las"=/las/->/[lah], as in Chile and Venezuela.
- /ɾ/: choose to represent /ɾ/ as /l/ in post-nuclear position (also known as lambdacism), which has been observed in some Caribbean dialects (Guitart, 1997), such as in Cuba and Puerto Rico (e.g. "porque"=/poɾke/->[polke]).

## 5. Methodology

The Spanish lexicon was extracted from two freely available sources. The first one is all the entries from the *Diccionario de la Real Academia de la Lengua Española* (REAL ACADEMIA ESPAÑOLA, 2021), and the text file was compiled by Domínguez (2015). The second database was available from the *LibreOffice* software (Foundation, T. D, 2020). This database contains word entries, ordered in alphabetic order. A great advantage of this resource is that there is a folder for each of the 16 countries available. This can be maximised in our research by selecting to use a pronunciation dictionary customised for a specific Spanish variant, following the dialectal features presented in the previous section. In our approach, we use countries as a proxy for dialectal variants, mainly in terms of orthographic spelling and words used in a specific country. The other way of using this is by merging two or more countries sources to encompass more accurate lexicon. For example, this can be used in the case where there is a study focusing on Caribbean Spanish, which would include Cuba, Colombia, Dominican Republic, Puerto Rico, Panama, and Venezuela (cf. Álvarez et al., 2009; Núñez-Méndez, 2021). In this case, users can choose these countries to create a single pronunciation dictionary.

The data processing was done in RStudio (R Core Team, 2021), by applying a wide range of scripts developed by the main author. The following section describes the segmentation and classification of the data, from words to their corresponding lexical and grammatical tagging.

### 5.1 Word Level

The phonology of Spanish allows an almost one to one correspondence when assigning individual letters to individual phonemes. This makes the computational process of annotation more reliable and accurate. This also considers the fact that in standard orthography, there are only two diagraphs in Spanish: "ch" (/tʃ/), and "ll" (/ʎ/ or /ʝ/). Another adjustment is the letter "h", which is silent in Spanish. Apart from these exceptions, the other letters can have a one-to-one representation between letters and phonemes. The main motivation for this approach is to allow the expansion of this dictionary and facilitate the automation of adding new entries, in which the algorithm can identify the letters and then create a pronunciation representation from there.

This process has its own challenges. The main one is when we create phonemic representations of borrowed words from another language. For example, the English word "today", following the automatic parsing would be 't o d a j', but this must be corrected to 't u d e j', if the aim is to reflect pronunciations that are closer to source language. In the current version, we have identified these words in a semi-automatic way, but we aim to fully automate this process in future versions. However, due to the open-source nature of the dictionary and its expandability option, this can be done by users as needed.

Another important aspect of standard Spanish orthography is the writing of the acute accent[6] in written words. This was maximised in our computational approach. The rules of orthography help identify where the stress is located within the syllable, which is the main function of the accent in Spanish[7]. However, not all stressed syllables are indicated with an accent. In those cases, these are exceptions when the accent is not written, and the stressed syllable can still be identified following the orthographic rules. This is expanded in Section 5.5.

---

[5] Here we do not make a clear-cut distinction between European and Latin-American Spanish. The first reason is because there are features that appear in some dialects within Spain that are also found in dialects of Latin-America. Also, Spanish in Latin-America is not homogenous since there are many distinct dialects, and characterising Latin-American Spanish as one single variant is inaccurate. However, with these features we aim to capture major dialectal distinctions, and new distinctions can be added to the dictionary.

[6] In this paper, we will use the word "accent" to refer to the *orthographic* marking in Spanish, and we will use the words "variant" and "dialect" to refer to the *speech-related* differences. We will also use the word "stress" for the *phonetic* emphasis of syllables within words.

[7] This is an important difference between accents in Spanish and other languages such as French. In French, for example, accents are used to indicate the nature of the vowel (open/close), whereas in Spanish is to indicate phonological stress.

Taking into consideration the orthographic rules, we then proceeded with the phonemic mapping. The first step was to break down each word into the individual letters. Each letter then was mapped to a pronunciation counterpart, which represents the phonemes. For this, all "h" letters were excluded since this affects the identification as shown below. For this mapping, we created representations that are one single letter per phoneme, which is similar to the CELEX annotation, with a combination of upper- and lower-case letters. Having one-to-one letter mapping facilitates analysis in further stages, for example, when counting characters, which is effective at dealing with transcriptions with no space between characters. Here, each individual letter would represent a single phoneme, thus the total number of characters would represent the total number of phonemes.

## 5.2 Vowel Classification

Vowels were classified into two groups: monophthongs or semivowels. If a vowel is not in contact with another vowel, then it was assigned its corresponding value as a monophthong. But if it was in contact with another vowel, it was classified as either a vowel or a semivowel. This depended on the stress pattern represented by the accent. If the vowel being classified did not carry the accent/stress, then it was assigned to the semivowel category. If it carried the accent, then it kept its monophthongal classification.

## 5.3 Consonant Classification

Our consonant classification process aimed to capture the main phonetic variations presented in Section 4. It does not mean to be exhaustive and further additions can be implemented in future versions of the tool. For the consonants, with only two exceptions, they were assigned their corresponding phonological representation following their orthography, for example, "p"='p', "m"='m'. The first exception was the letter "c". It can represent either a fricative (/s/ or /θ/) or a stop (/k/). If it was before "a, o, u", it was assigned 'k'. In the other cases it was either /s/ or /θ/, depending on the Spanish variant.

The second exception was on the voiced stops /b, d, g/. First, all contexts of "g" before "e, I", (similar to "c"), were assigned to 'h' or 'x' (depending on the variant), which is the case that before these vowel letters it is pronounced /h/ or /x/. Further, if "b, d, g" were between vowels, then they were converted to upper case 'B, D, G', respectively, to represent their intervocalic nature (lenited forms [β, ð, ɣ]).

## 5.4 Syllable Breakdown

The first step towards syllable breakdown was to identify the vowels in each word and then assign their onsets and codas. For onset and codas, Spanish allows up to two consonants in each cluster (Sheperd, 2003). In terms of the nucleus, at the phonetic level, a semivowel ([j, w]) can be before and after the nucleus, or both in the case of triphthongs. These rules follow the formula below.

$$O_{((C1)\ C2)}\ N_{(S)\ V\ (S)}\ C_{(C1\ (C2))}$$

For example, in the word "transporte", the vowel 'a' can take 'tr' as the onset. For the coda, the maximum it can have is two consonants, if it is the right cluster. In this case, 'ns' is a legal cluster. /p/ therefore starts the following syllable. In this syllable, even when it can take two consonants in the coda position, the cluster 'rt' is not a legal cluster in Spanish. Thus, the second syllable makes its boundary between 'r' and 't', leaving 't' as the onset of the last syllable. The final syllable breakdown is represented below:

trans|por|te
CCVCC|CVC|CV

## 5.5 Stress Assignment

The next step was to annotate the stressed syllable for each word. As mentioned before, Spanish orthography is an accurate guideline for making syllabic breakdown and stress patterns. The location of a stress is counted in Spanish from right to left, i.e. from the final syllable backwards. There are three positions a stress can take in word stems (Eddington, 2004): last syllable (*aguda*, e.g. "limón"), penultimate syllable (*grave*, e.g. "casa"), and antepenultimate (*esdrújula*, e.g. "búsqueda")[8].

In our process, the first step was to identify the location of an accent in the word. For those words that had the accents, the corresponding syllable was classified as stressed. If there was no accent, then it was identified following the rules of orthography. The first step was to identify whether the word ended in 'n', 's' or vowel. If this was the case, then the stress fell in the penultimate syllable. If the word ended in any other letter, then the stress was assigned to the last syllable.

## 5.6 Phonotactics and IPA Annotations

For the phonotactic labelling, we converted all vowels to V and all consonants to C. For the IPA notation, we converted each letter to their corresponding IPA representation. This was done with both syllabic divisions and without, and also by adding the order or segments for a more precise annotation. This allows to do searchers like: getting all consonants that appear as the second element of a consonant cluster (C1[C2]V).

## 5.7 Morphological and Lexical Annotation

This section describes the classification of all words based on their morphological and lexical information. For this, we implemented Natural Language Processing techniques in R through scripts developed by the main author. We used the UDPipe R package (Straka and Straková, 2017), and we used the GSD model for Spanish. This library is widely

---

[8] There are rules that include a fourth group: *sobre-esdrújulas*, which are argued that have the stress in the fourth or fifth syllable. However, these are generally a stem with suffixes. For example, "ágilmente" has two morphemes: "ágil" + "mente". This has an impact on the pronunciation pattern where the syllable "-men-" can have a secondary stress. Because of this, most of the literature agrees that there are only three places of stress in Spanish. For more on secondary stress see

used for tasks such as tokenization, tagging and lemmatization for many languages. We used this package to label all words in the dictionary. The distributions of the main lexical categories are presented in the figure below with the corresponding final counts across all the dictionary.

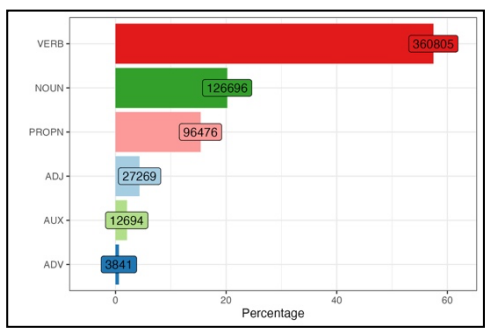

Figure 1: Distribution of Lexical Categories in the final dictionary, with their corresponding counts.

## 6. Results

All these steps and data processing stages, give a final dictionary with 628,300 entries, fully annotated and readily available for Spanish data research. The dictionary (ESPADA) can be accessed here:
https://github.com/simongonzalez/ESPADA

A sample of the entries and the columns is shown in Table 5 below.

| *Entry* | *POS* | *Base* | *Phonotactics* | *IPA* |
|---|---|---|---|---|
| *aarón* | PROPN | a r O n | V CVC | a ɾon |
| *con* | ADP | k O n | CVC | kon |
| *gris* | NOUN | g r I s | CCVC | gɾis |
| *la* | DET | l A | CV | la |
| *mesa* | NOUN | m E s a | CV CV | me sa |

Table 5: Sample Data from the final dictionary file.

The dictionary contains entries for 16 Spanish speaking countries. If only the entries for a given dialect are chosen, there is an average of 58,876 words per country, with varying total numbers, as shown in Figure 2. This gives users the option to select the dialect, or group of dialects, that best apply to the forced alignment process.

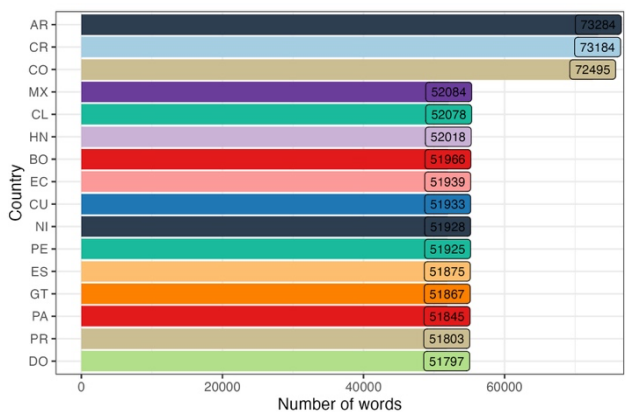

Figure 2: Word Distributions across Countries, with Argentina having the greatest number of entries and Dominican Republic with the least.

## 7. Conclusions

In this paper, we presented the development of a Spanish dictionary for phonetic and lexical tagging in socio-phonetic research. With this tool, researchers can have the maximum freedom to choose the dictionary that is the most representative of the data to be forced aligned, and then analysed in further stages. It has been our aim to facilitate the process of forced alignment and data wrangling for meaningful and accurate phonetic analysis. The open-source nature of this project also allows users to make the necessary changes to capture the complexity of phonetic variation in Spanish dialects.

## 8. Future Directions

This paper presents the development of the tool. Future work will aim to evaluate this dictionary with the other dictionaries compared in this paper. This will also include the assessment on the alignment accuracy across different Spanish dialects, especially in socio-phonetic studies.

## 9. Acknowledgements

We would like to acknowledge *The International Association for Forensic Phonetics and Acoustics (IAFPA)* for funding the main project encompassing this work. We give our thanks for their invaluable support. We also want to thank the two anonymous reviewers for their comments and suggestions. Their contribution has improved this paper in innumerable ways and saved us from many errors. Those that inevitably remain are entirely our own responsibility.

## 10. Bibliographical References